\documentclass[10pt,twocolumn,letterpaper]{article}

\usepackage{cvpr}
\usepackage{times}
\usepackage{epsfig}
\usepackage{graphicx}
\usepackage{amsmath}
\usepackage{amssymb}
\usepackage{multirow}
\usepackage{algorithm}
\usepackage{algorithmicx}
\usepackage{algpseudocode}
\newcommand{\mapAtoB}{G_{A\rightarrow B}}
\newcommand{\mapBtoA}{G_{B\rightarrow A}}
\usepackage[breaklinks=true,bookmarks=false]{hyperref}

\cvprfinalcopy % *** Uncomment this line for the final submission

\def\cvprPaperID{****} % *** Enter the CVPR Paper ID here

\begin{document}

%%%%%%%%% TITLE
\title{Conditional Image-to-Image Translation}

\author{First Author\\
Institution1\\
Institution1 address\\
{\tt\small firstauthor@i1.org}
\and
Second Author\\
Institution2\\
First line of institution2 address\\
{\tt\small secondauthor@i2.org}
}
\author{Jianxin Lin\textsuperscript{1}~~~~~~Yingce Xia\textsuperscript{1}~~~~~~Tao Qin\textsuperscript{2}~~~~~~Zhibo Chen\textsuperscript{1}~~~~~~Tie-Yan Liu\textsuperscript{2}\\
            \textsuperscript{1}University of Science and Technology of China~~~~~~\textsuperscript{2}Microsoft Research Asia\\
            \tt\small linjx@mail.ustc.edu.cn~~~~~~\tt\small yingce.xia@gmail.com\\
            \tt\small \{taoqin, tie-yan.liu\}@microsoft.com~~~~~~\tt\small chenzhibo@ustc.edu.cn\\
            }

\maketitle
\pagestyle{empty}  % no page number for the second and the later pages
\thispagestyle{empty} % no page number for the first page
%\thispagestyle{empty}

%%%%%%%%% ABSTRACT
\begin{abstract}
Image-to-image translation tasks have been widely investigated with Generative Adversarial Networks (GANs) and dual learning. However, existing models lack the ability to control the translated results in the target domain and their results usually lack of diversity in the sense that a fixed image usually leads to (almost) deterministic translation result. In this paper, we study a new problem, conditional image-to-image translation, which is to translate an image from the source domain to the target domain conditioned on a given image in the target domain. It requires that the generated image should inherit some domain-specific features of the conditional image from the target domain.  Therefore, changing the conditional image in the target domain will lead to diverse translation results for a fixed input image from the source domain, and therefore the conditional input image helps to control the translation results. We tackle this problem with unpaired data based on GANs and dual learning. We twist two conditional translation models (one translation from A domain to B domain, and the other one from B domain to A domain) together for inputs combination and reconstruction while preserving domain independent features. We carry out experiments on men's faces from-to women's faces translation and edges to shoes\&bags translations. The results demonstrate the effectiveness of our proposed method.
\end{abstract}

\section{Introduction}
Image-to-image translation covers a large variety of computer vision problems, including image stylization~\cite{gatys2016image}, segmentation~\cite{long2015fully} and saliency detection~\cite{goferman2012context}. It aims at learning a mapping that can convert an image from a source domain to a target domain, while preserving the main presentations of the input images. For example, in the aforementioned three tasks, an input image might be converted to a portrait similar to Van Gogh's styles, a heat map splitted into different regions, or a pencil sketch, while the edges and outlines remain unchanged. Since it is usually hard to collect a large amount of parallel data for such tasks, unsupervised learning algorithms have been widely adopted. Particularly, the generative adversarial networks (GAN)~\cite{goodfellow2014generative} and dual learning~\cite{he2016dual,pmlr-v70-xia17a} are extensively studied in image-to-image translations. \cite{Yi_2017_ICCV,kim2017learning,zhu2017unpaired} tackle image-to-image translation by the aforementioned two techniques, where the GANs are used to ensure the generated images belonging to the target domain, and dual learning can help improve image qualities by minimizing reconstruction loss.

An implicit assumption of image-to-image translation is that an image contains two kinds of features\footnote{ Note that the two kinds of features are relative concepts, and domain-specific features in one task might be domain-independent features in another task, depending on what domains one focuses on in the task.}: {\em domain-independent features}, which are preserved during the translation (i.e., the edges of face, eyes, nose and mouse while translating a man' face to a woman' face), and {\em domain-specific features}, which are changed during the translation (i.e., the color and style of the hair for face image translation).  Image-to-Image translation aims at transferring images from the source domain to the target domain by preserving domain-independent features while replacing domain-specific features.

While it is not difficult for existing image-to-image translation methods to convert an image from a source domain to a target domain, it is not easy for them to control or manipulate the style in fine granularity of the generated image in the target domain. Consider the gender transform problem studied  in~\cite{kim2017learning}, which is to translate a man's photo to a woman's. Can we translate Hillary's photo to a man' photo with the hair style and color of Trump? DiscoGAN~\cite{kim2017learning} can indeed output a woman's photo given a man's photo as input, but cannot control the hair style or color of the output image. DualGAN~\cite{Yi_2017_ICCV,zhu2017unpaired} cannot implement this kind of fine-granularity control neither.
\begin{figure}[!htbp]
\centering
\includegraphics[width=7.5cm]{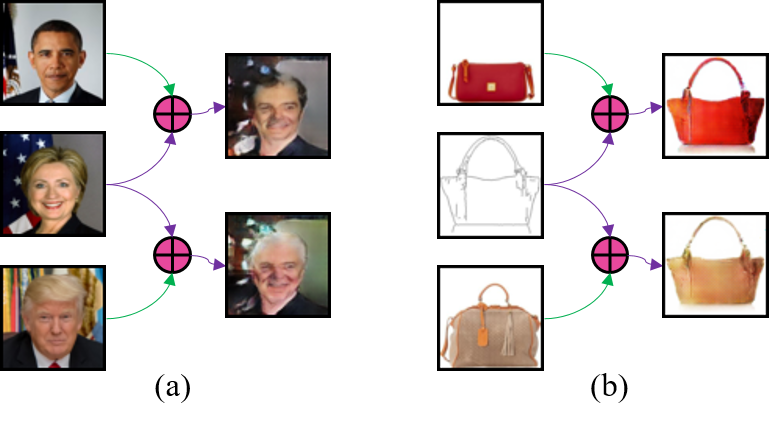}
	\caption{Conditional image-to-image translation. (a) Conditional women-to-men photo translation. (b) Conditional edges-to-handbags translation. The purple arrow represents translation flow and the green arrow represents the conditional information flow.}
	\label{fig:examples_in_intro}
\end{figure}
To fulfill such a blank in image translation, we propose the concept of {\em conditional image-to-image} translation, which can specify domain-specific features in the target domain, carried by another input image from the target domain. An example of conditional image-to-image translation is shown in Figure~\ref{fig:examples_in_intro}, in which we want to convert Hillary's photo to a man's photo. As shown in the figure, with an addition man's photo as input, we can control the translated image (e.g., the hair color and style).

\subsection{Problem Setup}
We first define some notations. Suppose there are two image domains $\mathcal{D}_A$ and $\mathcal{D}_B$.  Following the implicit assumption, an image $x_A\in\mathcal{D}_A$  can be represented as $x_A=x_A^i \oplus x_A^s$, where $x^i_A$'s are domain-independent features, $x^s_A$'s are domain-specific features, and $\oplus$ is the operator that can merge the two kinds of features into a complete image. Similarly, for an image $x_B\in\mathcal{D}_B$, we have  $x_B=x_B^i \oplus x_B^s$. Take the images in Figure~\ref{fig:examples_in_intro} as examples: (1) If the two domains are man's and woman's photos, the domain-independent features are individual facial organs like eyes and mouths and the domain-specific features are beard and hair style. (2) If the two domains are real bags and the edges of bags, the domain-independent features are exactly the edges of bags themselves, and the domain-specific are the colors and textures. %We can assume that there is no domain-specific features for the edges of bags.

%Please note that the shared features and the identity features we mentioned above are two relative concepts, depending on the two domains we focus on. That is, the shared features might be identity features in other tasks, and vice versa. For example, (1) in Figure~\ref{fig:examples_in_intro}(a), the colors can be seen as shared features, however, they are identity features in Figure~\ref{fig:examples_in_intro}(b); (2) the outlines of the instances in Figure~\ref{fig:examples_in_intro}(b) are shared features; but in Figure~\ref{fig:examples_in_intro}(a), they can be used to justify whether it is a man or a woman in a photo and thus, should be categorized as identity features.

The problem of conditional image-to-image translation from domain $\mathcal{D}_A$ to $\mathcal{D}_B$ is as follows: Taken an image $x_A\in\mathcal{D}_A$ as input and an image $x_B\in\mathcal{D}_B$ as conditional input, outputs an image $x_{AB}$ in domain $\mathcal{D}_B$ that keeping the domain-independent features of $x_A$ and combining the domain-specific features carried in $x_B$, i.e.,
\begin{equation}
\begin{aligned}
x_{AB} = \mapAtoB(x_A,x_B) = x_A^i \oplus x_B^s,
\end{aligned}
\label{eq:our_prime_task}
\end{equation}
where $\mapAtoB$ denotes the translation function. Similarly, we have the reverse conditional translation
\begin{equation}
\begin{aligned}
x_{BA} = \mapBtoA(x_B,x_A) = x_B^i \oplus x_A^s.
\end{aligned}
\label{eq:our_dual_task}
\end{equation}

For simplicity, we call $\mapAtoB$ the forward translation and $\mapBtoA$ the reverse translation.  In this work we study how to learn such two translations.

\subsection{Our Results}
There are three main challenges in solving the conditional image translation problem. The first one is how to extract the domain-independent and domain-specific features for a given image. The second is how to merge the features from two different domains into a natural image in the target domain. The third one is that there is no parallel data for us to learn such the mappings.

To tackle these challenges, we propose the {\em conditional dual-GAN} (briefly, cd-GAN), which can leverage the strengths of both GAN and dual learning. Under such a framework, the mappings of two directions, $\mapAtoB$ and $\mapBtoA$, are jointly learned. The model of cd-GAN follows the encoder-decoder based framework: the encoder is used to extract the domain-independent and domain-specific features and the decoder is to merge the two kinds of features to generate images. We chose GAN and dual learning due to the following considerations: (1) The dual learning framework can help learn to extract and merge the domain-specific and domain-independent features by minimizing carefully designed reconstruction errors, including reconstruction errors of the whole image, the domain-independent features, and the domain-specific features. (2) GAN can ensure that the generated images well mimic the natural images in the target domain. (3) Both dual learning \cite{he2016dual,Yi_2017_ICCV,zhu2017unpaired} and GAN \cite{goodfellow2014generative,radford2015unsupervised,denton2015deep} work well under unsupervised settings.

We carry out experiments on different tasks, including face-to-face translation, edge-to-shoe translation, and edge-to-handbag translation. The results demonstrate that our network can effectively translate image with conditional information and robust to various applications.

Our main contributions lie in two folds: (1) We define a new problem,  conditional image-to-image translation, which is a more general framework than conventional image translation. (2) We propose the cd-GAN algorithm to solve the problem in an end-to-end way.

The remaining parts are organized follows. We introduce related work in Section \ref{related work} and present the details of cd-GAN in Section \ref{framework}, including network architecture and the training algorithm. Then we report experimental results in Section \ref{experiment} and conclude in Section \ref{conclusion}.

\section{Related Work}\label{related work}

Image generation has been widely explored in recent years. Models based on variational autoencoder (VAE) \cite{kingma2013auto} aim to improve the quality and efficiency of image generation by learning an inference network. GANs \cite{goodfellow2014generative} were firstly proposed to generate images from random variables by a two-player minimax game. Researchers have been exploited the capability of GANs for various image generation tasks. \cite{denton2015deep} proposed to synthesize images at multiple resolutions with a Laplacian pyramid of adversarial generators and discriminators, and can condition on class labels for controllable generation. \cite{radford2015unsupervised} introduced a class of deep convolutional generative networks (DCGANs) for high-quality image generation and unsupervised image classification tasks.

Instead of learning to generate image samples from scratch (i.e., random vectors), the basic idea of image-to-image translation is to learn a parametric translation function that transforms an input image in a source domain to an image in a target domain. \cite{long2015fully} proposed a fully convolutional network (FCN) for image-to-segmentation translation. Pix2pix \cite{isola2016image} extended the basic FCN framework to other image-to-image translation tasks, including label-to-street scene and aerial-to-map. Meanwhile, pix2pix utilized adversarial training technique to ensure high-level domain similarity of the translation results.

The image-to-image models mentioned above require paired training data between the source and target domains. There is another line of works studying unpaired domain translation. Based on adversarial training,  \cite{dumoulin2016adversarially} and \cite{donahue2016adversarial} proposed algorithms to jointly learn to map latent space to data space and project the data space back to latent space. \cite{taigman2016unsupervised} presented a domain transfer network (DTN) for unsupervised cross-domain image generation employing a compound loss function including multiclass adversarial loss and $f$-constancy component, which could generate convincing novel images of previously unseen entities and preserve their identity. \cite{he2016dual} developed a dual learning mechanism which can enable a neural machine translation system to automatically learn from unlabeled data through a dual learning game. Following the idea of dual learning, DualGAN~\cite{Yi_2017_ICCV}, DiscoGAN~\cite{kim2017learning} and CycleGAN~\cite{zhu2017unpaired} were proposed to tackle the unpaired image translation problem by training two cross domain transfer GANs at the same time. \cite{Luo_2017_ICCV} proposed to utilize dual learning for semantic image segmentation. \cite{lu2017conditional} further proposed a conditional CycleGAN for face super-resolution by adding facial attributes obtained from human annotation. However, collecting a large amount of such human annotated data can be hard and expensive.

In this work, we study a new setting of image-to-image translation, in which we hope to control the generated images in fine granularity with unpaired data. We call such a new problem \emph{conditional image-to-image translation}.

\begin{figure*}[!htpb]
	\centering
	\includegraphics[width=0.9\linewidth]{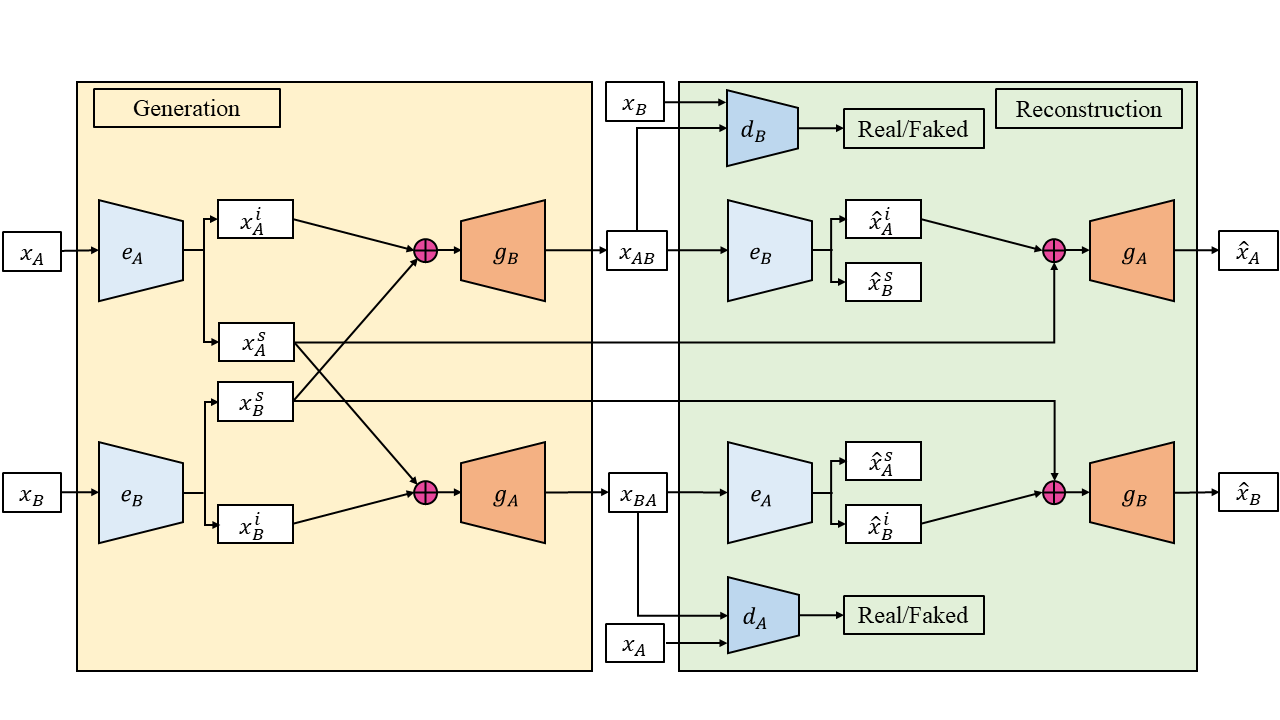}
	\caption{Architecture of the proposed conditional dual GAN (cd-GAN).}
	\label{fig:overall_framework}
\end{figure*}

\section{Conditional Dual GAN}\label{framework}
Figure \ref{fig:overall_framework} shows the overall architecture of the proposed model, in which the left part is an encoder-decoder based framework for image translation and the right part includes additional components introduced to train the encoder and decoder.

\subsection{The Encoder-Decoder Framework}\label{sec:network_arch}
As shown in the figure, there are two encoders $e_A$ and $e_B$ and two decoders $g_A$ and $g_B$.

The encoders serve as feature extractors, which take an image as input and output the two kinds of features, domain-independent features and domain-specific features, with the corresponding modules in the encoders. In particular, given two images $x_A$ and $x_B$, we have
\begin{equation}
(x_A^i, x_A^s) = e_A(x_A);\quad(x_B^i, x_B^s) = e_B(x_B).
\label{eq:encoder_func}
\end{equation}
If only looking at the encoder, there is no difference between the two kinds of features. It is the remaining parts of the overall model and the training process that differentiate the two kinds of features. More details are discussed in Section~\ref{sec:discussion_of_framework}.

The decoders serve as generators, which take as inputs the domain-independent features from the image in the source domain and the domain-specific features from the image in the target domain and output a generated image in the target domain.
That is,
\begin{equation}
x_{AB} = g_B(x_A^i, x_B^s);\quad x_{BA} = g_A(x_B^i, x_A^s).
\label{eq:generator_func}
\end{equation}

\subsection{Training Algorithm}\label{sec:training_procedure}
We leverage dual learning techniques and the GAN techniques to train the encoders and decoders. The optimization process is shown in the right part of Figure~\ref{fig:overall_framework}.

\subsubsection{GAN loss} To ensure the generated $x_{AB}$ and $x_{BA}$ are in the corresponding domains, we employ two discriminators $d_A$ and $d_B$ to differentiate the real images and synthetic ones. $d_A$ (or $d_B$) takes an image as input and outputs a probability  indicating how likely the input is a natural image from domain $\mathcal{D}_A$ (or $\mathcal{D}_B$). The objective function is
\begin{equation}
\begin{aligned}
\ell_{\text{GAN}}  =& \log(d_A(x_A)) + \log(1-d_A(x_{BA})) \\
                   +& \log(d_B(x_B)) + \log(1-d_B(x_{AB})).
\end{aligned}
\label{eq:gan_loss}
\end{equation}
The goal of the encoders and decoders $e_A$, $e_B$, $g_A$, $g_B$ is to generate images as similar to natural images and fool the discriminators $d_A$ and $d_B$, i.e., they try to minimize $\ell_{\text{GAN}}$. The goal of $d_A$ and $d_B$ is to differentiate generated images from natural images, i.e., they try to maximize $\ell_{\text{GAN}}$.

\subsubsection{Dual learning loss}
The key idea of dual learning is to improve the performance of a model by minimizing the reconstruction error. %Different from ~\cite{Yi_2017_ICCV,kim2017learning,zhu2017unpaired}, to translate an image to the target domain, we need to extract two kinds of features from the input images, due to which we cannot apply the same reconstruction loops in the previous literature.

To reconstruct the two images $\hat{x}_A$ and $\hat{x}_B$, as shown in Figure~\ref{fig:overall_framework}, we first extract the two kinds of features of the generated images:
\begin{equation}
(\hat{x}_A^i, \hat{x}_B^s) = e_B(x_{AB});\quad (\hat{x}_B^i, \hat{x}_A^s) = e_A(x_{BA}),
\label{eq:extract_reconstruct_feature}
\end{equation}
and then reconstruct images as follows:
\begin{equation}
\hat{x}_A = g_A(\hat{x}_A^i, x_A^s);\quad \hat{x}_B = g_B(\hat{x}_B^i, x_B^s).
\label{eq:reconstruct_images}
\end{equation}

We evaluate the reconstruction quality from three aspects: the image level reconstruction error $\ell_{\text{dual}}^{\text{im}}$, the reconstruction error $\ell_{\text{dual}}^{\text{di}}$ of the domain-independent features, and the reconstruction error $\ell_{\text{dual}}^{\text{ds}}$ of the domain-specific features as follows:
\begin{equation}
\ell_{\text{dual}}^{\text{im}}(x_A, x_B)= \Vert x_A - \hat{x}_A \Vert^2 + \Vert x_B - \hat{x}_B \Vert^2,
\label{eq:dual_im}
\end{equation}
\begin{equation}
\ell_{\text{dual}}^{\text{di}}(x_A, x_B)
= \Vert x_A^i - \hat{x}_A^i \Vert^2 + \Vert x_B^i - \hat{x}_B^i \Vert^2,
\label{eq:dual_di}
\end{equation}
\begin{equation}
\ell_{\text{dual}}^{\text{ds}}(x_A, x_B)
= \Vert x_A^s - \hat{x}_A^s \Vert^2 + \Vert x_B^s - \hat{x}_B^s \Vert^2.
\label{eq:dual_ds}
\end{equation}

Compared with the existing dual learning approaches~\cite{Yi_2017_ICCV} which only consider the image level reconstruction error, our method considers more aspects and therefore is expected to achieve better accuracy.

\subsubsection{Overall training process}
Since the discriminators only impact the GAN loss $\ell_{\text{GAN}}$, we only use this loss to compute the gradients and update $d_A$ and $d_B$. In contrast, the encoders and decoders impact all the 4 losses (i.e., the GAN loss and three reconstruction errors), we use all the 4 objectives to compute gradients and update models for them. Note that since the 4 objectives are of different magnitudes, their gradients may vary a lot in terms of magnitudes. To smooth the training process, we normalize the gradients so that their magnitudes are comparable across 4 losses.  We summarize the training process in Algorithm~\ref{alg_1}.

\begin{algorithm}
\caption{cd-GAN training process}
\label{alg_1}
\begin{algorithmic}[1]
\Require Training images $\{x_{A,i}\}_{i=1}^{m}\subset\mathcal{D}_A $, $\{x_{B,j}\}_{j=1}^{m}\subset\mathcal{D}_B$, batch size $K$, optimizer $Opt(\cdot,\cdot)$;
\State Randomly initialize $e_A$, $e_B$, $g_A$, $g_B$, $d_A$ and $d_B$.
%\Repeat
\State Randomly sample a minibatch of images and prepare the data pairs  $\mathcal{S}=\{(x_{A,k}, x_{B,k})\}^K_{k=1}$.
\State For any data pair $(x_{A,k},x_{B,k})\in\mathcal{S}$, generate conditional translations by Eqn.(\ref{eq:encoder_func},\ref{eq:generator_func}), and reconstruct the images by Eqn.(\ref{eq:extract_reconstruct_feature},\ref{eq:reconstruct_images});
\State Update the discriminators as follows:\newline
$d_A \leftarrow Opt(d_A, (1/K)\nabla_{d_A}\textstyle{\sum_{k=1}^{K}}\ell_{\text{GAN}}(x_{A,k}, x_{B,k}))$,\newline
$d_B \leftarrow Opt(d_B, (1/K)\nabla_{d_B}\textstyle{\sum_{k=1}^{K}}\ell_{\text{GAN}}( x_{A,k}, x_{B,k}))$;
\State For each $\Theta\in\{e_A, e_B, g_A, g_B\}$, compute the gradients \newline
$\Delta_{\text{GAN}} = (1/K)\nabla_{\Theta}\textstyle{\sum_{k=1}^{K}}\ell_{\text{GAN}}(x_{A,k}, x_{B,k})$,\newline
$\Delta_{\text{im}} = (1/K)\nabla_{\Theta}\textstyle{\sum_{k=1}^{K}}\ell_{\text{dual}}^{\text{im}}(x_{A,k}, x_{B,k})$,\newline
$\Delta_{\text{di}} = (1/K)\nabla_{\Theta}\textstyle{\sum_{k=1}^{K}}\ell_{\text{dual}}^{\text{di}}(x_{A,k}, x_{B,k})$,\newline
$\Delta_{\text{ds}} = (1/K)\nabla_{\Theta}\textstyle{\sum_{k=1}^{K}}\ell_{\text{dual}}^{\text{ds}}(x_{A,k}, x_{B,k})$,\newline
normalize the four gradients to make their magnitudes comparable, sum them to obtain $\Delta$, and $\Theta\rightarrow Opt(\Theta, \Delta)$.
\State Repeat step 2 to step 6 until convergence
%\Until{convergence}
\end{algorithmic}
\end{algorithm}

In Algorithm~\ref{alg_1}, the choice of optimizers $Opt(\cdot,\cdot)$ is quite flexible, whose two inputs are the parameters to be optimized and the corresponding gradients. One can choose different optimizers (e.g. Adam~\cite{kingma2014adam}, or nesterov gradient descend~\cite{nesterov1983method}) for different tasks, depending on common practice for specific tasks and personal preferences. Besides, the $e_A$, $e_B$, $g_A$, $g_B$, $d_A$, $d_B$ might refer to either the models themselves, or their parameters, depending on the context.

\subsection{Discussions}\label{sec:discussion_of_framework}
Our proposed framework can learn to separate the domain-independent features and domain-specific features. In Figure \ref{fig:overall_framework}, consider the path of $x_A\rightarrow e_A\rightarrow x_A^i\rightarrow g_B\rightarrow x_{AB}$. Note that after training we ensure that $x_{AB}$ is an image in domain $\mathcal{D}_B$ and the features $x_A^i$ are still preserved in $x_{AB}$. Thus, $x_A^i$ should try to inherent the features that are independent to domain $\mathcal{D}_A$. Given that $x_A^i$ is domain independent, it is $x_B^s$ that carries information about domain $\mathcal{D}_B$. Thus, $x_B^s$ is domain-specific features. Similarly, we can see that $x_A^s$ is domain-specific and $x_B^i$ is domain-independent.

DualGAN~\cite{Yi_2017_ICCV}, DiscoGAN~\cite{kim2017learning} and CycleGAN~\cite{zhu2017unpaired} can be treated as simplified versions of our cd-GAN, by removing the domain-specific features. For example, in CycleGAN, given an $x_A\in\mathcal{D}_A$, any $x_{AB}\in\mathcal{D}_B$ is a legal translation, no matter what $x_{B}\in\mathcal{D}_B$ is. In our work, we require that the generated images should match the inputs from two domains, which is more difficult.

Furthermore, cd-GAN works for both symmetric translations and asymmetric translations. In symmetric translations, both directions of translations need conditional inputs (illustrated in Figure \ref{fig:examples_in_intro}(a)). In asymmetric translations, only one direction of translation needs a conditional image as input (illustrated in Figure \ref{fig:examples_in_intro}(b)). That is, the translation from bag to edge does not need another edge image as input; even given an additional edge image as the conditional input, it does not change or help to control the translation result.

For asymmetric translations, we only need to slightly modify objectives for cd-GAN training.
Suppose the translation direction of $\mapBtoA$ does not need conditional input. Then we do not need to reconstruct the domain-specific features $x_A^s$. Accordingly, we modify the error of domain-specific features as follows, and other 3 losses do not change.
\begin{equation}
\ell_{\text{dual}}^{\text{ds}}(x_A, x_B)= \Vert x_B^s - \hat{x}_B^s \Vert^2
\label{eq:dual_dsa}
\end{equation}

\section{Experiments}\label{experiment}
We conduct a set of experiments to test the proposed model. We first describe experimental settings, and then report results for both symmetric translations and asymmetric translations. Finally we study individual components and loss functions of the proposed model.

\subsection{Settings}
For all experiments, the networks take images of $64\times 64$ resolution as inputs. The encoders $e_{A}$ and $e_{B}$ start with $3$ convolutional layers, each convolutional layer followed by leaky rectified linear units (Leaky ReLU) \cite{maas2013rectifier}. Then the network is splitted into two branches: in one branch, a convolutional layer is attached to extract domain-independent features; in the other branch, two fully-connected layers are attached to extract domain-specific features. Decoder networks $g_{A}$ and $g_{B}$ contain $4$ deconvolutional layers with ReLU units \cite{nair2010rectified}, except for the last layer using $\tanh$ activation function. The discriminators $d_A$ and $d_B$ consist of $4$ convolution layers, two fully-connected layers. Each layer is followed by Leaky ReLU units except for the last layer using sigmoid activation function. Details (e.g., number and size of filters, number of nodes in fully-connected layers) can be found in the supplementary document.

We use Adam~\cite{kingma2014adam} as the optimization algorithm with learning rate $0.0002$. Batch normalization is applied to all convolution layers and deconvolution layers except for the first and last ones. Minibatch size is fixed as $200$ for all the tasks.

We implement three related baselines for comparison.
\begin{enumerate}
\item\emph{DualGAN}~\cite{Yi_2017_ICCV,kim2017learning,zhu2017unpaired}. DualGAN was primitively proposed for unconditional image-to-image translation which does not require conditional input. Similar to our cd-GAN, DualGAN trains two translation models jointly.
\item \emph{DualGAN-c}. In order to enable DualGAN to utilize conditional input, we design a network as DualGAN-c. The main difference between DualGAN and DualGAN-c is that DualGAN-c translates the target outputs as Eqn.(\ref{eq:encoder_func},\ref{eq:generator_func}), and reconstructs inputs as $\hat{x}_A = g_A(e_B(x_{AB}))$ and $\hat{x}_B = g_B(e_A(x_{BA}))$.
\item \emph{GAN-c}. To verify the effectiveness of dual learning, we remove the dual learning losses of cd-GAN during training and obtain GAN-c.
\end{enumerate}

For symmetric translations, we carry out experiments on men-to-women face translations. We use the CelebA dataset~\cite{liu2015faceattributes}, which consists of $84434$ men's images (denoted as domain $\mathcal{D}_A$) and $118165$ women's images (denoted as domain $\mathcal{D}_B$). We randomly choose $4732$ men's images and $6379$ women's images for testing, and use the rest for training. In this task, the domain-independent features are organs (e.g., eyes, nose, mouse) and domain-specific features refer to hair-style, beard, the usage of lipstick.  For asymmetric translations, we work on edges-to-shoes and edges-to-bags translations with datasets used in \cite{yu2014fine} and \cite{zhu2016generative} respectively. In these two tasks, the domain-independent features are edges and domain-specific features are colors, textures, etc.
\begin{figure}[!htpb]
\centerline{\includegraphics[width=7.5cm]{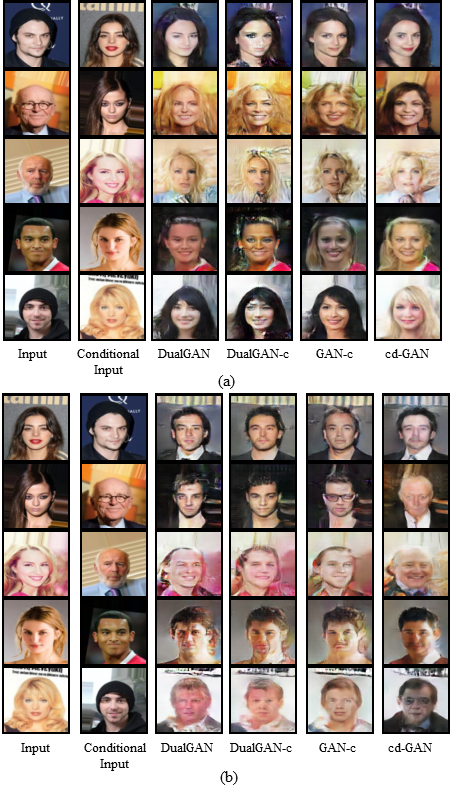}}
\caption{Conditional face-to-face translation. (a) Results of conditional men$\rightarrow$women translation. (b) Results of conditional women$\rightarrow$men translation.}
	\centering
	\label{fig:fig5}
\end{figure}

\subsection{Results}
\begin{figure}[!t]
	\centerline{\includegraphics[width=7.5cm]{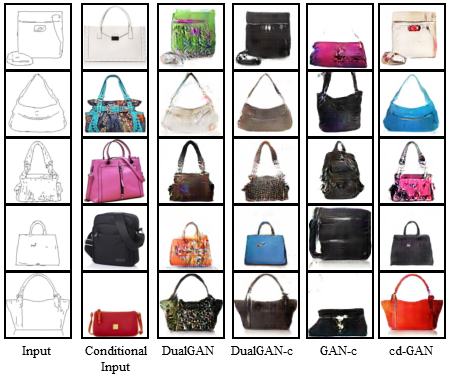}}
	\caption{Results of conditional edges$\rightarrow$handbags translation.}
	\centering
	\label{fig:fig6}
\end{figure}
\begin{figure}[!htpb]
	\centerline{\includegraphics[width=7.5cm]{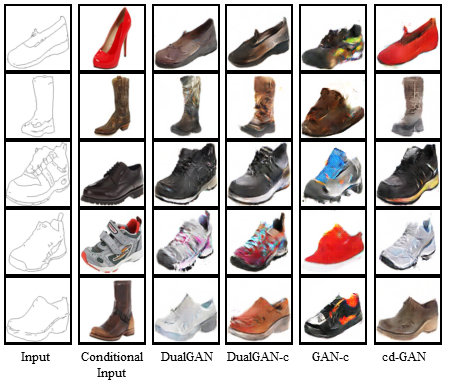}}
	\caption{Results of conditional edges$\rightarrow$shoes translation.}
	\centering
	\label{fig:fig7}
\end{figure}
The translation results of face-to-face, edges-to-bags and edges-to-shoes are shown in Figure~\ref{fig:fig5}-\ref{fig:fig7} respectively.

For men-to-women translations, from Figure~\ref{fig:fig5}(a), we have several observations. (1) DualGAN can indeed generate woman's photo, but its results are purely based on the men's photos, since it does not take the conditional images as inputs. (2) Although taking the conditional image as input, DualGAN-c fails to integrate the information (e.g., style) from the conditional input into its translation output. (3) For GAN-c, sometimes its translation result is not relevant to the original source-domain input, e.g., the 4-th row Figure~\ref{fig:fig5}(a). This is because in training it is required to generate a target-domain image, but its output is not required to be similar (in certain aspects) to the original input.  (4) cd-GAN works best among all the models by preserving domain-independent features from the source-domain input and combining the domain-specific features from the target-domain conditional input. Here are two examples. (1) In 6-th column of 1-st row, the woman is put on red lipstick. (2) In 6-th column of 5-th row, the hair-style of the generated image is the most similar to the conditional input.

We can get similar observations for women-to-men translations as shown in Figure~\ref{fig:fig5}(b), especially for the domain-specific features such as hair style and beard.

From Figure~\ref{fig:fig6} and ~\ref{fig:fig7}, we find that cd-GAN can well leverage the domain-specific information carried in the conditional inputs and control the generated target-domain images accordingly. DualGAN, DuanGAN-c and GAN-c do not effectively utilize the conditional inputs.

\begin{figure}
	\centerline{\includegraphics[width=8.5cm]{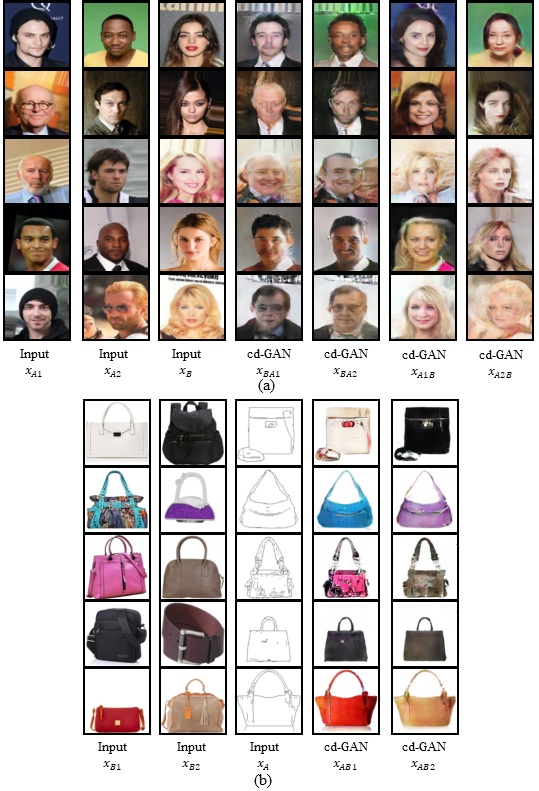}}
	\caption{Our cd-GAN model can produce diverse results with different conditional images. (a) Results of women$\rightarrow$men translation with two different men's images as conditional inputs. (b) Results of edges$\rightarrow$handbags translation with two different handbags as conditional inputs. }
	\centering
	\label{fig:fig8}
\end{figure}

One important characteristic of conditional image-to-image translation model is that it can generate diverse target-domain images for a fixed source-domain image, only if different target-domain images are provided as inputs. To verify such this ability of cd-GAN, we conduct two experiments: (1) for each woman's photo, we work on women-to-men translations with different man's photos as conditional inputs; (2) for each edge of a bag, we work on edges-to-bags translations with different bags as conditional inputs. The results are shown in Figure~\ref{fig:fig8}. Figure~\ref{fig:fig8}(b) shows that cd-GAN can fulfill edges with the colors and textures provided by the conditional inputs. Besides, cd-GAN also achieves reasonable improvements on most face translations: The domain-independent features like woman's facial outline, orientations and expressions are preserved, while the women specific features like hair-style and the usage the lipstick are replaced with men's. An example is the second row of Figure~\ref{fig:fig8}(a), where pointed chins, serious expressions and looking forward are preserved in the generated images. The hairstyles (bald v.s. short hair) and the beard (no beard v.s. short beard) are reflected by the corresponding men's. Similar translations of the other images can also be found. Note that there are several failure cases in face translations, such as first column of Figure \ref{fig:fig8} (a) and last column of Figure \ref{fig:fig8} (b). Most translated results demonstrate the effectiveness of our model. More examples can be found in our supplementary document.

%Since we have shown that our proposed cd-GAN is effective to transfer images with conditional information in the above subsections, we would like to further verify the variation of generated results when given various conditional information. As shown in Figure \ref{fig:fig8}, we illustrate our model's ability to produce various results on men faces to women faces translation and edges to handbags translation tasks when given different conditional images. We model also fails to absorb the conditional information in some cases, such as first column of Figure \ref{fig:fig8} (a) and last column of Figure \ref{fig:fig8} (b), most of the translated results still can demonstrate our model's effectiveness.

\subsection{Component Study}
\begin{figure}
  \centerline{\includegraphics[width=8.5cm]{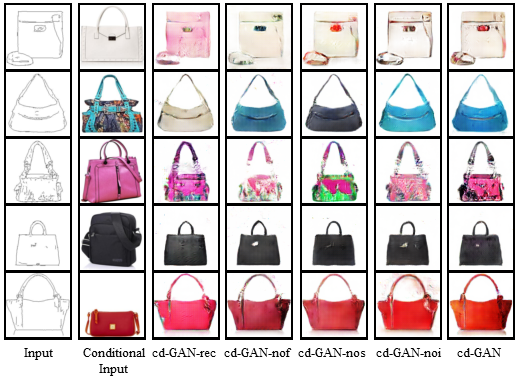}}
  \caption{Results produced by different connections and losses of cd-GANs. }
  \centering
\label{fig:fig9}
\end{figure}
In this sub section, we study other possible design choices for the model architecture in Figure \ref{fig:overall_framework} and losses used in training. We compare cd-GAN with other four models as follows:
\begin{itemize}
  \item \textbf{cd-GAN-rec}. The inputs are reconstructed as
  \begin{equation}
{\hat{x}_A} = g_A(\hat{x}_A^i,\hat{x}_A^s); {\hat{x}_B} = g_B(\hat{x}_B^i,\hat{x}_B^s)
\label{eq:dual_dsa222}
\end{equation}
  instead of Eqn.(\ref{eq:reconstruct_images}). That is, the connection from $x_A^s$ to $g_A$ in the right box of Figure \ref{fig:overall_framework} is replaced by the connection from $\hat{x}_A^s$ to $g_A$, and the connection from $x_B^s$ to $g_B$ in the right box of Figure \ref{fig:overall_framework} is replaced by the connection from $\hat{x}_B^s$ to $g_B$.
  \item \textbf{cd-GAN-nof}. Both domain-specific and domain-independent feature reconstruction losses, i.e., Eqn.\eqref{eq:dual_ds} and Eqn.\eqref{eq:dual_di}, are removed from dual learning losses.
  \item \textbf{cd-GAN-nos}. The domain-specific feature reconstruction loss, i.e., Eqn.\eqref{eq:dual_ds}, is removed from dual learning losses.
  \item \textbf{cd-GAN-noi}. The domain-independent feature reconstruction loss, i.e., Eqn.\eqref{eq:dual_ds} is removed from dual learning losses.
\end{itemize}
The comparison experiments are conducted on the edges-to-handbags task. The results are shown in Figure \ref{fig:fig9}. Our cd-GAN outperforms the other four candidate models with better color schemes. Failure of cd-GAN-rec demonstrates the necessity of ``skip connections'' (i.e., the connections from $x_A^s$ to $g_A$ and from $x_B^s$ to $g_B$) for image reconstruction. Since the domain-specific feature level and image level reconstruction losses have implicitly put constrains on domain-specific feature to some extent, the results produced by cd-GAN-noi are closest to results of cd-GAN among the four candidate models.
%\subsection{Influence of Main Hyper-parameters}
%Our cd-GAN mainly consists of three hyper-parameters, i.e., $\lambda,\alpha,\beta$. In our implementation, we set $\lambda=1$ and $\beta=1$ to all conditional translation tasks, while we set the different $\alpha$ according to different tasks since different tasks may require different ability of domain-specific feature constrain. To investigate the influence of hyper-parameter $\alpha$, we set the different value of $\alpha$ for edges-to-handbags task. The results is shown in Figure \ref{fig:fig10}. As we can see, although there remains some difference caused by different $\alpha$, the main style of these results is quite similar, which indicates that our cd-GAN is able to generate results in a wide range of $\alpha$.
%\begin{figure}
%  \centerline{\includegraphics[width=8.5cm]{figure/fig10}}
%  \caption{Different results produced by different values of $\alpha$ of cd-GANs. }
%  \centering
%\label{fig:fig10}
%\end{figure}

\begin{figure}
  \centerline{\includegraphics[width=7.5cm]{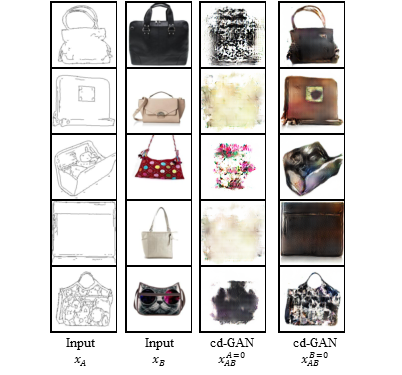}}
  \caption{Images generated using only domain-independent features or domain-specific features.  }
  \centering
\label{fig:feature_visual}
\end{figure}

So far, we have shown the translation results of cd-GAN generated from the combination domain-specific features and domain-independent features. One may be interested in what we really learn in the two kinds of features. Here we try to understand them by generating translation results using each kind of features separately:
\begin{itemize}
\item We generate an image using the domain-specific features only:
$$x_{AB}^{A=0} = g_B(x_A^i=\mathbf{0}, x_B^s),$$ in which we set the domain-independent features to 0.
\item  We generate an image using the domain-independent features only:
$$x_{AB}^{B=0} = g_B(x_A^i, x_B^s=\mathbf{0}),$$  in which we set the domain-specific features to 0.
\end{itemize}
The results are shown in Figure \ref{fig:feature_visual}. As we can see, the image $x_{AB}^{A=0}$ has similar style to $x_B$, which indicates that our cd-GAN can indeed extract domain-specific features. While $x_{AB}^{B=0}$ already loses conditional information of $x_B$, it still preserves main shape of $x_A$, which demonstrates that cd-GAN indeed extracts domain-independent features.
\subsection{User Study}
\begin{figure}[!htbp]
\centering
\includegraphics[width=6.5cm]{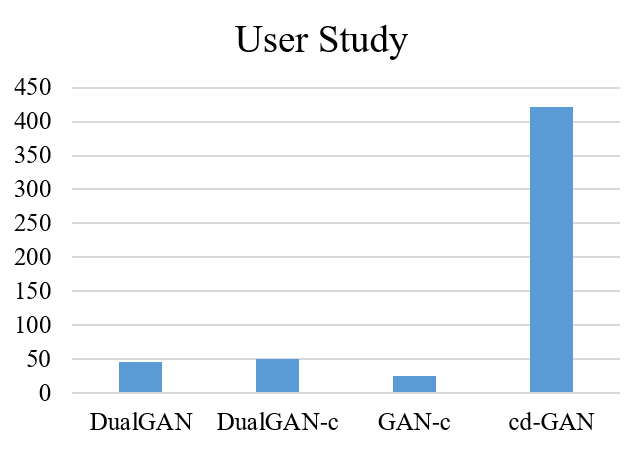}
	\caption{The result of user study.}
	\label{fig:user_study}
\end{figure}
We have conducted user study to compare domain-specific features similarity between generated images and conditional images. Total $17$ subjects ($10$ males, $7$ females, age range $20-35$) from different backgrounds are asked to make comparison of $32$ sets of images. We show the subjects source image, conditional image, our result and results from other methods. Then each subject selects generated image most similar to conditional image. The result of user study shows that our model obviously outperforms other methods.
\section{Conclusions and Future Work}\label{conclusion}
In this paper, we have studied the problem of conditional image-to-image translation, in which we translate an image from a source domain to a target domain conditioned on another target-domain image as input. We have proposed a new model based on GANs and dual learning. The model can well leverage the conditional inputs to control and diversify the translation results. Experiments on two settings (symmetric translations and asymmetric translations) and three tasks (face-to-face, edges-to-shoes and edges-to-handbags translations) have demonstrated the effectiveness of the proposed model.

There are multiple aspects to explore for conditional image translation. First, we will apply the proposed model to more image translation tasks. Second, it is interesting to design better models for this translation problem. Third, the problem of conditional translations may be extend to other applications, such as conditional video translations and conditional text translations.
\section{Acknowledgement}
This work was supported in part by the National Key Research and Development Program of China under Grant No$. 2016$YFC$0801001$, NSFC under Grant $61571413$, $61632001$, $61390514$, and Intel ICRI MNC.

{\small
\bibliographystyle{ieee}
\bibliography{Bibliography-File}
}

\end{document}